# Attribute CNNs for Word Spotting in Handwritten Documents

Sebastian Sudholt · Gernot A. Fink

**Abstract** Word spotting has become a field of strong research interest in document image analysis over the last years. Recently, AttributeSVMs were proposed which predict a binary attribute representation [3]. At their time, this influential method defined the state-of-the-art in segmentation-based word spotting. In this work, we present an approach for learning attribute representations with Convolutional Neural Networks (CNNs). By taking a probabilistic perspective on training CNNs, we derive two different loss functions for binary and real-valued word string embeddings. In addition, we propose two different CNN architectures, specifically designed for word spotting. These architectures are able to be trained in an end-to-end fashion. In a number of experiments, we investigate the influence of different word string embeddings and optimization strategies. We show our Attribute CNNs to achieve state-of-the-art results for segmentation-based word spotting on a large variety of data sets.

## 1 Introduction

Understanding the contents of handwritten texts from document images has long been a traditional field of research in computer science. Despite its long history, it's still considered an unsolved task as classification systems are still not able to consistently achieve results as are common for machine printed text recognition. This is especially the case when the text to be recognized either exposes a large amount of degradation or if the variability in the word images of the same class is high. In these situations, using a retrieval instead of a recognition approach produces more robust results. This retrieval approach has been termed *Keyword Spotting* or simply *Word Spotting*. Here, the database consists of document or word images.

There exists a variety of different query types with *Query-by-Example (QbE)* and *Query-by-String (QbS)* being the most prominent ones. In QbE applications, the query is a word image whereas in QbS it is a textual string representation. With respect to practical applications, QbE poses certain limitations as the user has to identify a query word image from a document image collection. This might either already solve the task ("does the collection contain the query?") or be tedious when looking for infrequent words as queries [2,48].

Thus more recently, the focus has shifted towards QbS-based approaches [2,3,43]. One notable drawback of this method, however, is that the word spotting system has to learn a mapping from textual to visual representation first. Most of the time, this can only be achieved through manually annotated training samples.

An elegant solution for enabling a word spotting system to perform QbE as well as QbS are common subspace approaches. Here, the textual representation and the word image representation are projected into a common subspace in which the word spotting task boils down to a simple nearest neighbor search. A very successful approach in this regard has been the *embedded attributes* framework [3]. The projection for the text is done by computing binary textual attributes in

Sebastian Sudholt
TU Dortmund University
E-mail: sebastian.sudholt@tu-dortmund.de

Gernot A. Fink
TU Dortmund University
E-mail: gernot.fink@tu-dortmund.de

a $d$-dimensional space. Each attribute then represents one dimension in a common attribute space. This attribute representation is called *Pyramidal Histogram of Characters (PHOC)*. The projection from word images to common subspace is then learned by an ensemble of attribute detectors. More specifically, the images are encoded into a Fisher Vector representation which is then forwarded to $d$ Support Vector Machines (SVM), each predicting one attribute of the PHOC. This ensemble of SVMs is referred to as *AttributeSVMs*.

The approach of [3] has certain design aspects that can be improved upon. First, the feature representation and the attribute detectors (SVMs) are optimized separately. Moreover, the individual SVMs each have to learn their own model and make no use of shared parameters. As there exist strong correlations between certain attributes of the PHOC, parameter sharing could help an attribute detector in terms of training time as well as detection accuracy.

In this paper we present an approach to word spotting by using Convolutional Neural Networks (CNN) which are able to predict multiple attributes at the same time. The CNNs optimize the feature representation and the attribute detectors in a combined, supervised fashion which leads to discriminative features and highly accurate representations. By leveraging attributes, the CNNs are able to predict representations for word image classes with high precision, even if they were not present at training time (out of vocabulary). The presented CNNs are capable of dealing with binary as well as real-valued attributes. In the fashion of AttributeSVMs, we refer to our CNNs as *Attribute CNNs*. Fig. 1 gives an overview of how we use Attribute CNNs in order to perform word spotting.

The contributions of this paper are as follows: We design two CNNs for word spotting which are able to predict binary as well as real-valued attribute representations. For this, we present a theoretical framework which allows for designing loss functions and also interpret the output and training of a CNN from a theoretical point of view. This framework is not only applicable to our problem at hand but can be used for any task where CNNs are to be employed. Furthermore, we investigate the relationship between two common loss functions for learning real-valued representations, namely the Cosine Loss and the Euclidean Loss. In addition to building our Attribute CNNs from well-known CNN layers, we propose a novel pooling layer called *Temporal Pyramid Pooling layer*. This layer is especially suitable for processing word images of varying width and height. Finally, we evaluate our method on a total of six publicly available data sets featuring both Latin and Arabic script. The results show that our CNNs are able to achieve equal or better results than the current state-of-the-art. While the presented Attribute CNNs are designed specifically for word spotting applications, they could principally be used in any attribute classification scenario.

Preliminary versions and results of the presented approach have already been published in ICFHR 2016 [58] and ICDAR 2017 [59]. While [58] covers the initial ideas of training a CNN with an attribute representation, [59] introduces the TPP layer and reports further experiments. In addition, [59] also introduces the Cosine Loss for attribute CNNs but without giving theoretical explanations as to why this loss is suitable for the task at hand. This aspect is addressed specifically in this work.

## 2 Related Work

2.1 Deep Learning and CNNs

The recent success of *Convolutional Neural Networks (CNNs)* has been sparked by a number of developments which enabled these neural networks to become the backbone of state-of-the-art *deep learning* systems. Key contributions in this regard have been the introduction of *Rectified Linear Units (ReLU)* as activation function [14], highly optimized implementations running on graphics cards [19], specialized weight initialization strategies [16] and large scale data sets such as ImageNet [49] to train on.

The architecture of classic CNNs like LeNet [30], AlexNet [26] or VGG16 [53] all share a convolutional part in the early layers and a set of fully connected layers at the very end of the neural network. More recent CNN designs like GoogLeNet [61], the All Convolutional Net [55] or Residual Networks [17] ignore the fully connected layers in large parts and build up the individual CNN of (almost) only convolutional layers.

2.2 Attribute Representations

The typical classification problem in computer vision deals with assigning one out of $d$ labels to a given image. However, classification systems can, in general, only assign class labels that have been seen during training. In order to alleviate this problem, semantic attributes were proposed [9,27,28]. Instead of a single class label, a number of attributes can be assigned to a class. Each class is thus represented by a specific set of attributes while single attributes are shared by all classes. If a classifier is trained to predict the attributes, knowledge about these semantic units can then be transferred from the training to the test classes even if there exist classes which were not seen during training [27]. If there attribute configuration is known, they can be classified even if no samples were seen during training.

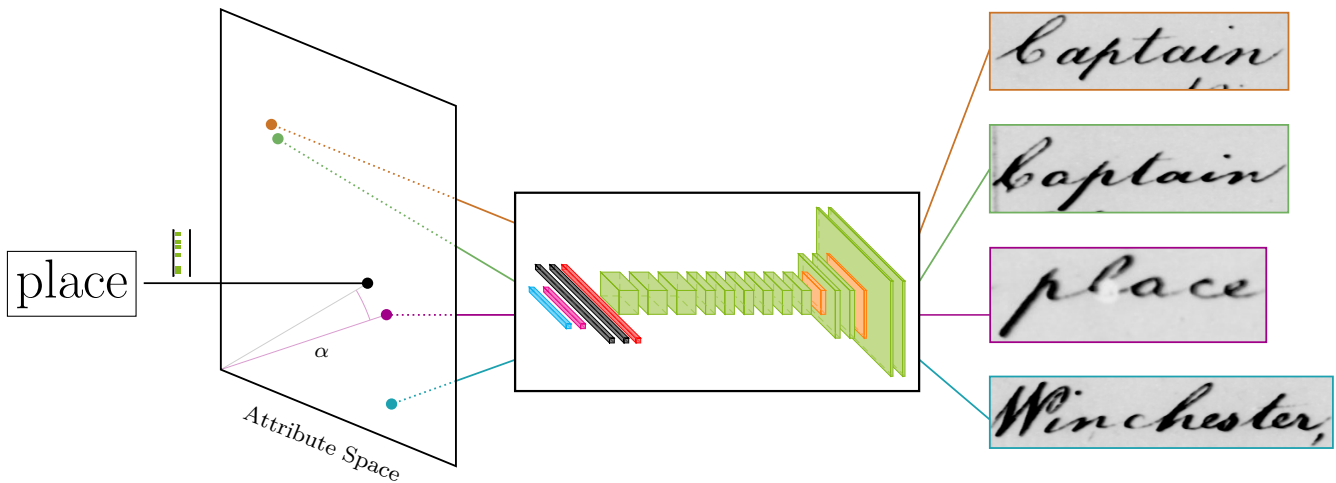

Fig. 1: The figure visualizes the proposed system. For Query-by-String, the attribute representation is extracted directly from the string (left side). For word images, a CNN predicts the attribute representation. This way, annotations and word images can be projected into a joint attribute space. Similarity in the attribute space is determined by applying the cosine distance, i.e., by determining the angle between attribute vectors. Not shown in the figure is Query-by-Example, which is done by simply ranking the attribute representation obtained from the word images in a nearest neighbor approach.

2.3 Word Spotting

In the following, related work with respect to word spotting is presented. For space limitations, we only give a brief overview of this field of research. For a detailed survey on word spotting see [13].

The goal of word spotting is to extract relevant parts of document image collections with respect to a certain query. The primary applications are browsing and indexing document image collections, especially in situations where a recognition, i.e., transcription approach does not achieve satisfying results [38,44,47]. However, word spotting can be used for other tasks as well such as refining OCR results [52].

In [31] word spotting was first applied to handwritten documents which is generally considered a much harder task compared to doing word spotting on machine printed document images due to large variations in writing style. While [31] used XOR-maps on binary features, ensuing works often made use of sequential methods which had proven successful in the field of word image recognition. The two most prominent among these are *Dynamic Time Warping (DTW)* [23,38] or *Hidden Markov Models (HMM)* [10]. Sequential models are still employed in recent approaches in the form of *Semi-continuous HMMs* [40], *Bag-of-Features HMMs (BoF-HMM)* [44] and *Bidirectional Long Short Term Memory (BLSTM)* networks [11]. However, there has been an ever growing amount of work focusing on holistic representations as well. In [2,57] and [47] densely extracted SIFT descriptors are used in a Bag-of-Features approach. The quantized descriptors are aggregated into a *Spatial Pyramid* [29] to form a holistic word image descriptor which can then be used in a simple nearest neighbor approach for word spotting.

With the exception of QbE, one has to find a model to map from the query representation to the word image. In [43] this is achieved through a BoF-HMM. Other approaches make use of label embedding techniques. In [2] the holistic word image descriptor and a $n$-gram-based textual descriptor are merged together and projected into a subspace. In a different and very influential approach, *embedded attributes* are used as common representations [3]. This method allows for an efficient framework, in which QbS as well as QbE can be performed. For this, the transcription of a word image is mapped to a binary attribute representation called *Pyramidal Histogram of Characters (PHOC)*. Extracting a PHOC from a given transcription is visualized in Fig. 2. The first level encodes which characters of the given alphabet are present in the entire word. Note that if a character appears multiple times it is not counted but simply denoted as present. For the second layer, the PHOC encodes whether characters are present in the left and right split of the transcription. For determining whether a character is present in a split or not, the normalized occupancy $Occ(k, n) = \left[\frac{k}{n}, \frac{k+1}{n}\right]$ is used [3]. Here, $n$ is the length of the transcription and $k$ is the position index of a specific character in the word. If the normalized occupancy overlaps at least 50% with a given split the character is considered present in this split. For example, the character $a$ is present in both splits in the second level in Fig. 2 as the occupancy overlaps 50% with both partitions. A nice trait of the PHOC representation is that it can be extracted from a given transcription directly without the need for any sort of training.

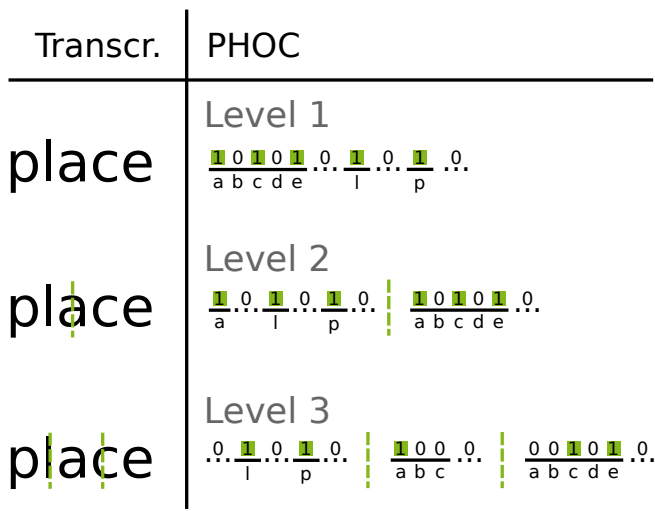

Fig. 2: The figure shows how to extract a 3-level PHOC from a given transcription. At each level, the word string is split into a certain number of sections. For each section, the presence or absence of characters from a given alphabet is determined and saved in a binary histogram. The resulting histograms are concatenated in order to form the PHOC vector.

In order to predict PHOCs from a given word image, [3] makes use of an ensemble of SVMs which are trained on the Fisher Vector representation of the word image. Each SVM predicts one attribute in the PHOC. This ensemble is referred to as *AttributeSVMs* by the authors [3].

## 2.4 Word Spotting with CNNs

Due to their recent success in other fields of computer vision, CNNs have been increasingly used for word spotting as well. One of the first works in this regard was presented in [51]. Here, the authors finetuned an AlexNet pretrained on the ImageNet database to predict word image classes. The CNN features from the second to last layer are then used in order to perform word spotting. The results presented are already quite competitive. However, the framework does not allow for QbS word spotting.

Simultaneously to the preliminary conference version [58] of this paper, three other approaches investigated the use of attribute representations in combination with CNNs. In [36], an approach very close to ours is used in a word recognition task. Here, the authors make use of a custom architecture which processes fixed sized word images and outputs a PHOC representation. In their approach, they extend the PHOC representation in order to account for one level of trigrams. Each level of the PHOC is then predicted by an individual MLP. All MLPs, however, make use of a shared convolutional part of the network.

The approach in [24] also makes use of a CNN in order to learn PHOCs, albeit not in an end-to-end fashion as is done in our case. The authors pretrain a network architecture inspired by [18] on a synthetically generated data set of one million word images [25]. Then they fine-tune on the training partition of the respective data sets used in their evaluation. For predicting attributes, they take outputs of certain layers from the neural network and use them as features for training an AttributeSVMs. It is unclear, however, which layer is used for feature extraction. This approach is very similar to the one presented by [3] in that [24] simply replace the Fisher Vectors with features from their CNN.

A different approach is presented by [64] in form of a Triplet-CNN. Here, a Residual Network [17] is used in combination with a Soft Positive Negative Triplet Loss [4] in order to learn holistic descriptors for word images. These descriptors are then used as features for training an MLP to predict the desired attribute representation. For this they make use of the the Cosine Embedding Loss. Using this special loss function, not only binary attribute representations can be used for training but also real-valued label vectors. In addition to PHOC vectors, the authors investigate their approach using a new word string embedding method called Discrete Cosine Transform of Words (DCToW) which achieves comparable results to the PHOC representation. For the DCToW, an indicator matrix is build which uses a given alphabet as row and the letter positions in the word as column indices. In each column, the respective character position from the alphabet is marked with a 1. Afterwards, a DCT is applied to each row of he matrix. The largest three coefficients from each row are then concatenated in order to form the final DCToW descriptor.

In our previous work of this paper, we have also investigated the use of the *Spatial Pyramid of Characters (SPOC)* descriptor as attribute representation for word images. The SPOC is very similar to the PHOC. However, instead of denoting character presence or absence with a binary value in each level, the SPOC creates histograms of characters, i.e., counts the occurrence of each character in each level.

## 3 Attribute CNNs

In this section we describe our proposed Attribute CNN architectures. The two important aspects which we elaborate on are the loss functions used and the architectures themselves. First, we explain how the loss functions for our Attribute CNNs are derived. The concepts explained with respect to the loss functions are of a general nature and do not only apply for word spotting or document image analysis applications. Second, we present our Attribute CNNs and explain certain design choices.

### 3.1 Loss Functions for Attribute CNNs

Traditionally, CNNs have been heavily used in the domain of multi-class recognition. The task here is to predict one out of $k$ classes for a given image. Usually, this is achieved by computing the element-wise softmax

$$\hat{y}_i(\mathbf{o}) = \frac{e^{o_i}}{\sum_{j=1}^{d} e^{o_j}} \quad (1)$$

for each element $i$ in the last layer's output $\mathbf{o}$ which represents the posterior probability for the $i$-th of $d$ classes. The predicted class is the one with highest probability. For training, a one-hot-encoded vector is supplied to the CNN with the sought-after class having a numerical value of 1 and all other classes a value of 0. When dealing with attribute representations, the softmax output is infeasable as at most one element in the output vector can become 1. Attribute representations, however, are often times made up of a number of binary (e.g. PHOC) or real-valued (e.g. DCToW, SPOC) labels. In order to get the output for binary targets in the correct range one could use the sigmoid function as activation in the last layer. This leads to the question what loss function is suitable for learning such a representation. A straightforward approach would be to make use of the Euclidean Loss

$$l_\mathcal{N} = \frac{1}{2} \sum_{i=1}^{n} ||\mathbf{y} - \hat{\mathbf{y}}||_2^2. \quad (2)$$

This, however, bares the drawback, that the overall gradient is scaled by the derivative of the sigmoid in the last layer [33]. If the initialization of the networks weights leaves the sigmoid neurons in a saturated state (large or small values before the activation) this causes a slow convergence behavior or even a complete stall in training. The sigmoid activation function could, of course, be replaced by a linear function. This way, however, the output of the CNN would not be bounded anymore and the CNN could produce output values outside of the desired range. Whether a sigmoid or a linear activation is used, there exists another disadvantage: By using $l_\mathcal{N}$ one implicitly assumes that the Euclidean distance is a feasible metric for comparing vectors (in our case binary or real-valued attribute representations). As for high-dimensional vectors the ratio of closest and farthest points approaches 1 [1,7], the Euclidean distance is not a suitable metric in our situation.

It is obvious, that the activation function in the last layer and the loss function are tightly coupled when training neural networks. A very elegant framework for finding suitable combinations of activation functions and loss functions are *Generalized Linear Models (GLMs)*. Moreover, GLMs also allow for interpreting the training of neural networks from a probabilistic perspective. In the following, we derive loss functions and corresponding activation functions for binary as well as real-valued representations from these statistical models.

#### 3.1.1 Binary Attribute Representations

GLMs are statistical tools for predicting the expected value of a random variable $Y$ conditioned on an independent random variable $X$, e.g., [50, p. 281]. When using a GLM, two important assumptions are made: First, $Y$ follows a distribution from the exponential family and the expected value $E[Y]$ depends on a transformation of a linear combination of $X$. For the linear combination, the GLM uses a so-called *linear predictor*:

$$\eta(\mathbf{x}) = \mathbf{w}^T \mathbf{x}, \quad (3)$$

where $\mathbf{x}$ is a realization of $X$ and $\mathbf{w}$ are the parameters or weights of the GLM. The output of the GLM is then forwarded to a so-called *link function* which maps the linear prediction into a suitable range. The concept of GLMs is demonstrated in the following example: Assume we want to predict a single binary attribute from a word image. It can be reasonably assumed that $Y$, i.e., the random variable from which we assume the labels are drawn, follows a Bernoulli distribution. The probability mass function of a Bernoulli distributed variable

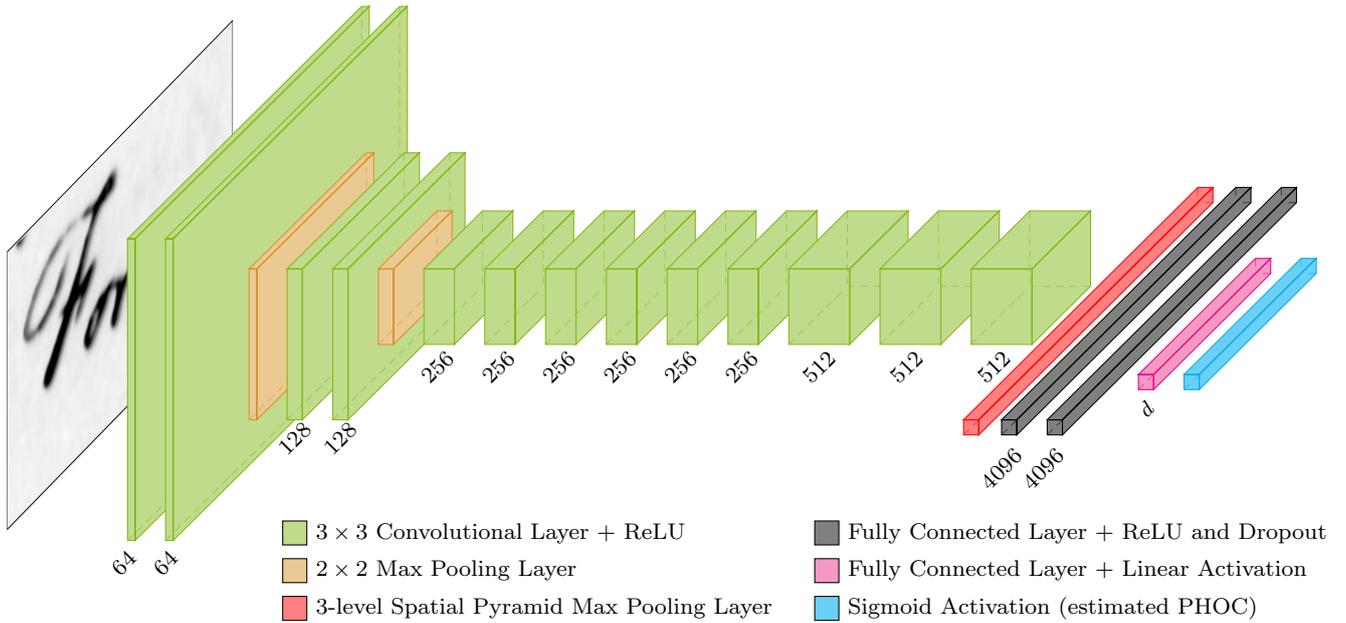

Fig. 3: Visualization of the PHOCNet architecture. All convolutional layers make use of $3 \times 3$ convolutions and are followed by a ReLU activation. Each convolutional layer applies a padding of 1 pixel on each side to its input feature map in order to preserve dimensionalities. The pooling layers downsample the input feature maps by applying max pooling over a $2 \times 2$ region with a step size of 2. A 50% dropout is applied to the output of the first two fully connected layers (black) during training.

is defined as

$$f_\mathcal{B}(k,p) = p^k (1-p)^{1-k} \text{ for } k \in \{0,1\} \quad (4)$$

where $p$ is the probability of drawing a 1 from $Y$ and $k$ is an indicator variable for the desired event. The GLM predicts the conditional expected value $\mathbb{E}[Y \mid X = \mathbf{x}]$ for the dependent variable $Y$. As the expected value for any Bernoulli distributed variable is again the probability $p$, the GLM thus predicts the posterior probability for $Y = 1$ given $\mathbf{x}$. We will denote this prediction of the posterior probability as $\hat{y}$. In the case of a Bernoulli distributed dependent variable, the link function of choice is the sigmoid function

$$\sigma(x) = \frac{1}{1+\exp(-x)} \quad (5)$$

where $x$ is a generic argument to the sigmoid function. The sigmoid squashes the result from the linear predictor to $(0,1)$ thus putting the output of the GLM in the correct range for a Bernoulli distributed variable.

Putting it all together, the GLM $g_\mathcal{B}$ for a Bernoulli distributed random variable predicts the conditional probability $\hat{y}$ through

$$\hat{y} = g_\mathcal{B}(\mathbf{x}, \mathbf{w}) = \frac{1}{1+\exp(-\mathbf{w}^T \mathbf{x})}. \quad (6)$$

This special case of using a Bernoulli distributed random variable $Y$ is also known as *logistic regression*. As can be seen from Eq. 6, the logistic regression can also be interpreted as a single layer, fully-connected neural network with sigmoid activation (similar to a perceptron). As is the case for neural networks, the GLM is trained by tuning the weights $\mathbf{w}$ using a training set $S = \{(\mathbf{x}^{(1)}, y^{(1)}), \ldots, (\mathbf{x}^{(n)}, y^{(n)})\}$ of samples $\mathbf{x}^{(i)}$ and their annotation $y^{(i)}$. Training is performed through *Maximum Likelihood Estimation (MLE)*: The negative log-likelihood function for a Bernoulli distributed variable is given by

$$l_\mathcal{B}(\mathbf{w} \mid S) = \sum_{i=1}^{n} -\log f_\mathcal{B}\left(y^{(i)}, g_\mathcal{B}\left(\mathbf{x}^{(i)}, \mathbf{w}\right)\right). \quad (7)$$

The estimated model parameters are obtained by minimizing the negative log-likelihood with respect to the weights of the GLM. Here, the dependent variable $Y$ represents the label distribution.

As hinted at earlier, the GLM can be compared to a single-layer neural network. This model can of course simply be replaced by switching the linear predictor (Eq. 3) with a deep neural network without invalidating any of the previous statements. The only difference is that instead of being able to solve the MLE analytically, it has to be done in the standard backpropagation framework with the negative log-likelihood serving as loss function. This allows us to create problem-specific loss functions by simply assuming a suitable probability distribution for the dependent variable $Y$, i.e., the label distribution. For example, substituting Eq. 4 into Eq. 7 yields the following loss function for Bernoulli distributed variables:

$$l_{\mathcal{B}} = -\sum_{i=1}^{n} y^{(i)} \log \hat{y}^{(i)} + \left(1 - y^{(i)}\right) \log \left(1 - \hat{y}^{(i)}\right). \quad (8)$$

Here, $\hat{y}^{(i)}$ is the output of the neural network for the $i$-th sample and $y^{(i)}$ the label. It can be shown, that minimizing $l_{\mathcal{B}}$ is equivalent to minimizing the cross entropy between the output distribution of the network and the label distribution [33].

To sum it all up: Training a neural network with sigmoid activation functions in the last layer using the Binary Cross Entropy Loss is equivalent to maximum likelihood estimation of the weights given the training set and assuming that the predicted variable follows a Bernoulli distribution. The output of the network can directly be interpreted as posterior probability for the attribute being 1 given an input sample $\mathbf{x}$.

Up to this point, we have only considered a single Bernoulli distributed variable as output of the network. In the case of binary attribute representations such as the PHOC, however, a neural network has to deal with a number of binary attributes. A straightforward approach would thus be to assume that $Y$ follows a multivariate Bernoulli distribution [6]. This approach has the drawback that the neural network used would have to have an output layer size of $2^d$ where $d$ is the dimensionality of the attribute representation. In our experiments, typical PHOC sizes are larger than 540 which would demand an output layer size greater than $3.599 \cdot 10^{162}$.

In order to make the problem tractable, the assumption can be made that the attribute representation is a collection of $d$ independent and Bernoulli distributed variables, each having their own probability $p$ of evaluating to 1. This way, we can compute $d$ separate loss functions and simply add up their values to form the final loss:

$$l_{\mathcal{B}} = -\sum_{i=1}^{n} \sum_{j=1}^{d} y_j^{(i)} \log \hat{y}_j^{(i)} + \left(1 - y_j^{(i)}\right) \log \left(1 - \hat{y}_j^{(i)}\right). \quad (9)$$

This generalization of the loss function for single Bernoulli distributed variables is known as *Binary Cross Entropy Loss*. Due to the squashing function used, another common name is *Sigmoid Cross Entropy Loss*.

*3.1.2 Real-valued Attribute Representations*

When dealing with real valued representations, we can again make use of the GLM framework and adapt it to account for the different data characteristics. A straightforward approach would be to assume that the label $\mathbf{y}$ is a set of $d$ random variables, each following a Normal distribution, similar to the assumption made for $l_{\mathcal{B}}$ (Eq. 9). However, applying MLE this leads to the Euclidean loss $l_{\mathcal{N}}$ (Eq. 2) with all the drawbacks mentioned above.

When dealing with high-dimensional representations, the cosine distance

$$d_{\cos}(\mathbf{a}, \mathbf{b}) = 1 - \frac{\mathbf{a}^T \mathbf{b}}{||\mathbf{a}|| \cdot ||\mathbf{b}||} \quad (10)$$

has proven effective in a number of applications for computing similarity between two vectors $\mathbf{a}$ and $\mathbf{b}$, e.g., [2, 3, 47]. In order to transfer this to the GLM framework, a distribution for $Y$ would be desirable that depends on the angle between vectors.

A very prominent distribution in this regard is the von Mises-Fisher distribution. Its probability density function for a $d$-dimensional vector is defined as

$$f_{\mathcal{MF}}(\mathbf{x}, \boldsymbol{\mu}, \kappa) = C_d(\kappa) \exp\left(\kappa \boldsymbol{\mu}^T \mathbf{x}\right) \quad (11)$$

where $\boldsymbol{\mu}$ is the mean direction, $\kappa$ is the concentration parameter and $C$ is a normalization constant, depending on the dimensionality of the data and $\kappa$. $\boldsymbol{\mu}$ and $\mathbf{x}$ are required to have unit length. The von Mises-Fisher distribution can be considered as a normal distribution on a $d$-dimensional hypersphere with the mean direction as analogy to the mean and the concentration parameter as analogy to the (inverse) variance. The density value for a given sample, however, depends on the angle of the sample to the mean direction and not its Euclidean distance. Another property of the von Mises-Fisher distribution is that it belongs to the exponential family thus making it suitable for the GLM framework. As link function, we choose the normalization function. This leads to the following GLM model:

$$\hat{\mathbf{y}} = g_{\mathcal{MF}}(\mathbf{x}, \mathbf{W}) = \frac{\mathbf{W}\mathbf{x}}{||\mathbf{W}\mathbf{x}||_2}. \quad (12)$$

The negative log-likelihood function of the model given a training data set $S$ is then defined by

$$l_{\mathcal{MF}}(\mathbf{W} \mid S) = \sum_{i=1}^{n} -\log f_{\mathcal{MF}}\left(\mathbf{x}^{(i)}, g_{\mathcal{MF}}\left(\mathbf{x}^{(i)}, \mathbf{W}\right), \kappa\right) \quad (13)$$

$$= \sum_{i=1}^{n} 1 - \cos\left(\mathbf{y}^{(i)}, \hat{\mathbf{y}}^{(i)}\right). \quad (14)$$

Again, this function can be used as loss function for training a neural network. The loss is thus simply the Cosine distance between the prediction and the label (Eq. 10) which is why it is known as *Cosine Loss*. Although this loss in itself is not novel, e.g., [5], this is, to the best of our knowledge, the first time it has been theoretically motivated from a statistical point of view. This motivation helps in understanding the assumptions which are made when using the Cosine Loss for training.

Interestingly, the Cosine Loss and the Euclidean Loss (Eq. 2) are equal given that both the labels $\mathbf{y}^{(i)}$ and the outputs $\hat{\mathbf{y}}^{(i)}$ of the network are normalized:

$$\frac{1}{2}\sum_{i=1}^{n} ||\mathbf{y} - \hat{\mathbf{y}}||_2^2 = \sum_{i=1}^{n} \frac{1}{2}\left(\mathbf{y}^T\mathbf{y} - 2\mathbf{y}^T\hat{\mathbf{y}} + \hat{\mathbf{y}}^T\hat{\mathbf{y}}\right)$$
$$= \sum_{i=1}^{n} \frac{1}{2}\left(2 - 2\mathbf{y}^T\hat{\mathbf{y}}\right)$$
$$= \sum_{i=1}^{n} 1 - \cos(\mathbf{y}, \hat{\mathbf{y}})$$

3.2 Attribute CNN Architectures for Word Spotting

The above-mentioned loss functions could simply be attached to any existing CNN architecture in order for the network to predict attributes. However, most CNN architectures have been designed with the task of operating on natural images in mind. The problem in this work is concerned with document or word images which often times have different properties than can be found in typical natural images. Hence, we specifically design a set of CNN architectures suitable to be applied to document images.

*3.2.1 PHOCNet*

As the first proposed architecture was originally designed to predict PHOCs [58], we dubbed it *PHOCNet*. The architecture is visualized in figure Fig. 3. It is inspired by the successful VGG16 architecture [53]. Just as in [53], we use a small number of filters in the lower convolutional layers and increase the amount of filters in the higher convolutional layers. This forces the CNN to learn less and thus more general features in the first layers while giving it the possibility to learn a large number of more abstract features later in the architecture. Additionally, we only use $3 \times 3$ filters in all convolutional layers. This imposes a regularization on the filters kernels [53].

The previous two design choices work for both natural images in the case of the VGG16 as well as word images in our case. However, there are certain aspects to be taken into account when designing a CNN architecture for document image analysis applications. In our case, we want the CNN to predict a representation for previously segmented word images. Usually, these word images exhibit a large variation in size. In related natural image applications such as multi-class classification, size variations are combated by either anisotropically rescaling an image to a fixed size [26] or sampling different crops from the image [17,53]. In the case of segmentation-based word spotting, we would like to obtain a holistic representation of a word image. We could possibly rescale all input images but this leads to severe distortions whenever the original aspect ratio does not approximately match the desired aspect ratio. Likewise, cropping is infeasable in our application as well: When predicting PHOCs for different crops of the input image, it is unclear how the final representations should be merged. While in multi-class classification problems the outputs for different crops can simply be averaged, this is not possible in our case as PHOCs contain a certain level of positional encoding of the attributes. Fusing outputs for different crops is thus very cumbersome and not straightforward at all.

In order to be able to process input images of different sizes, we employ a Spatial Pyramid Pooling (SPP) layer after the last convolutional layer (red layer in Fig. 3) [15]. This way, the first fully connected layer following the convolutional part is always presented with a fixed size image representation, independent of the input image size.

As we do not alter the input image sizes, we only use two pooling layers in the PHOCNet architecture. This way, we can even process the smallest word images in the data sets we tested on ($26 \times 26$ pixels for the George Washington data set). The pooling layers are placed rather close to the input layer in order to lower the computational cost (orange layers in Fig. 3).

*3.2.2 TPP-PHOCNet*

The use of the SPP layer enables the PHOCNet to accept almost arbitrarily sized input images and still output a representation of constant size. The SPP layer layout follows the one used for the "classic" spatial pyramid [29]: the number of cells along each dimension in a layer is doubled with respect to the previous layer.

In the field of document image analysis, however, using this layout was found to lead to inferior results compared to other cell partitions when using spatial pyramids on top of Bag-of-Feature representations [2,46,47].

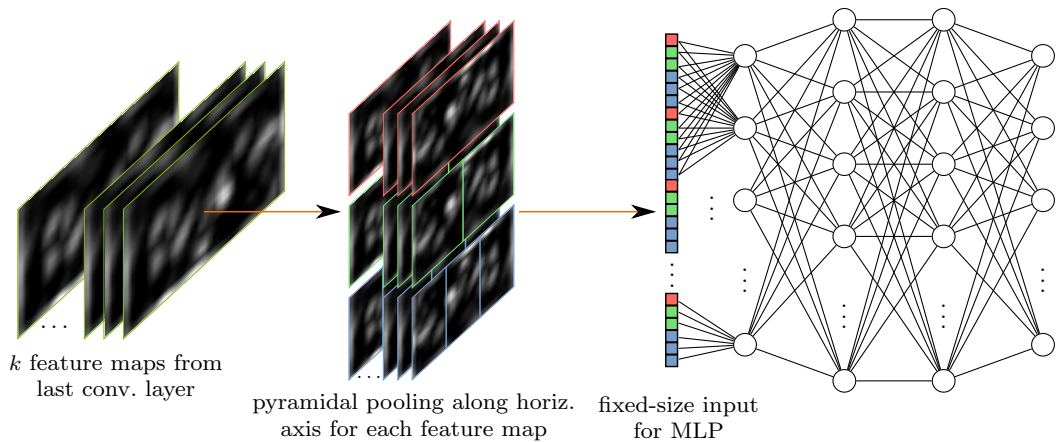

Fig. 4: The figure visualizes the TPP Layer. For every feature map of the last convolution layer (left) a sequence of max pooled values is extracted in a pyramidal fashion (middle). Here, a TPP Layer of size 1, 2, 3 is visualized. The layer produces output values for 9 max pooled regions per feature map. These values are stacked like in an SPP Layer in order to form a representation of fixed size for variably sized input feature maps. This representation is then fed to the MLP-part of the network (right).

Here, the retrieval results can be increased when using spatial pyramids featuring a fine-grained split along the horizontal axis while only using a rough partitioning for the vertical axis of a word image. This concept is pursued even further by HMM-based approaches, such as SC-HMMs [40], Bag-of-Feature HMMs [43, 44] or HMMs using word graphs [63], as well as methods based on Recurrent Neural Networks such as BLSTMs [11]. In the case of these sequential models, the vertical axis is not partitioned at all while the partitioning of the horizontal axis is implicitly done by splitting the word image in frames and processing them as sequence. In a way, this can be seen as a probabilistic version of a spatial pyramid.

In general, choosing a fine grained cell partitioning along the horizontal axis and a coarser partitioning along the vertical axis is important when dealing with word images. Incorporating this observation into a neural network layer, we propose a modified version of the SPP layer which we term *Temporal Pyramid Pooling (TPP) layer*. Pooling in this layer is done similarly to the PHOC pooling of binary attributes: Each layer splits the entire image into $n$ horizontal cells where $n$ is the index of the layer. Each cell covers the entire vertical axis of the word image. The pooling is thus only done along the axis of writing and each cell roughly represents features from consecutive intervals of the word image. When stacking multiple of these pooling layers with different amounts of splits along the axis of writing, we end up with a pyramidal representation encoding the progression of writing, hence the name Temporal Pyramid Pooling. This concept is visualized in Fig. 4. Here, the feature maps of the last convolution layer (left part of the figure) are followed by the temporal pyramid pooling approach described above.

For evaluating the TPP layer, we simply swap the SPP layer with the TPP layer in our experiments. The rest of the PHOCNet architecture is left unchanged. In our experiments, we use a TPP layer with max pooling and levels 1, 2, 3, 4 and 5 where each level indicates the amount of pooling bins along the horizontal axis. With the last convolution layer having 512 filters, the output size of the TPP Layer amounts to 7680.

3.3 Word Spotting with Attribute Representations

All our Attribute CNNs predict an attribute representation for a given word image. In order to perform word spotting or retrieval in general with these networks, we need a ranking functionality that is able to compare these representations. In other word spotting applications that make use of holistic word image representation, this has been achieved through a simple nearest neighbor approach with a suitable metric, e.g., [2, 3, 47]. The most widely used metric in word spotting is the Cosine distance (Eq. 10) which has been exclusively used in the literature when dealing with PHOC representations [3, 24, 36, 64]. Therefore, we adopt this metric for our word spotting method as well.

## 4 Experimental Evaluation

4.1 Data Sets

We evaluate our approach on six publicly available data sets in order to assess the performance of our Attribute CNNs and compare our approach to recent state-of-the-art methods from the literature.

*4.1.1 George Washington*

The *George Washington database (GW)* has become the standard benchmark for word spotting. It consists of 20 pages of correspondences between George Washington and his associates dating from 1755. It is an excerpt of a larger collection available at the library of congress[1]. As the documents in the George Washington data set are obtained from the letter book 2, which is not an original, but a later re-copied volume, it can be assumed that the data set has been produced by a single writer only.

There actually exist two versions of this data set which have been used to evaluate word spotting methods. The first version contains binarized word images which have been slant-corrected[2]. The second version[3] contains the plain gray-level document images and is by far the one more commonly used for evaluating word spotting methods, e.g., in [3, 44, 46–48, 51]. For our experiments, we will make use of the plain gray-level document images as well.

The annotation contains 4860 segmented words with 1124 different transcriptions [47]. As there exist no official training and test partitions, we follow the approach proposed in [3] and perform a fourfold cross validation. In order to be able to compare our results to those in [3], we use the exact same cross validation splits[4].

*4.1.2 IAM DB*

Although designed for handwriting recognition, the *IAM Database* [32] has become a major word spotting benchmark as well. It consists of more than 13 000 text lines containing a total of more than 115 000 words. The official partitioning splits the database in 6161 lines for training, 1840 for validation and 1861 for testing. One of the main challenges of this data set is that each writer

Table 1: Training set sizes for the Botany and Konzilsprotokolle data sets

|           | Botany | Konzilsprotokolle |
|-----------|--------|-------------------|
| Train I   | 1684   | 1849              |
| Train II  | 5295   | 7817              |
| Train III | 21 981 | 16 919            |

contributed to only one partition (either training, validation or test).

*4.1.3 Esposalles*

The *ESPOSALLES Database* [42] is an excerpt of a larger collection of marriage license books at the archives of the Cathedral of Barcelona. Among the major difficulties of the data set are several forms of degradation such as uneven illumination, smearing or bleed-through as well as high variability in script.

We use the official training and test partitioning that comes with the database. Overall, the annotation contains 32 052 training word images and 13 048 test word images.

*4.1.4 IFN/ENIT*

The *IFN/ENIT database* [35] contrasts all other data sets used as it features Arabic script. It consists of word images of Tunisian town or village names. The total amount of word images is 26 459 which were contributed by 411 different writers. There exists an official partitioning of the database into four subsets A, B, C and D. As is custom in handwriting recognition benchmarks run on the IFN/ENIT database, we use sets A, B and C for training and D for testing.

In order to extract attribute representations from the Arabic annotation we use a reduced character set which is generated in the following way: First all character shapes are mapped to their representative Arabic characters. Characters with optional Shadda diacritic are replaced with characters without the Shadda diacritic. Special two-character-shape ligature models are mapped to two-character ligature models without the shape contexts. This mapping produces a character set of size 50 for this data set.

*4.1.5 Botany and Konzilsprotokolle*

The two data sets *Botany in British India* and *Alvermann Konzilsprotokolle* where introduced and used in the Handwritten Keyword Spotting Competition held during the 2016 International Conference on Frontiers in Handwriting Recognition[5]. While the former covers botanical topics such as gardens, botanical collection and useful plants, the latter is a collection of protocols from the central administration at the university library of Greifswald, Germany, dating from 1794 to 1797.

---

[1] https://memory.loc.gov/ammem/gwhtml/
[2] http://www.fki.inf.unibe.ch/databases/iam-historical-document-database/washington-database
[3] http://ciir.cs.umass.edu/downloads/old/data_sets.html
[4] cross validation partitions available at https://github.com/almazan/watts/tree/master/data

As part of the competition was to evaluate how well the participating systems deal with small to large training data sets, each data set comes with three increasingly larger sets for training. Tab. 1 lists the sizes for the three different sets. The respective test set sizes are 3230 for Botany and 3533 for Konzilsprotokolle.

Different from the other data sets, Botany and Konzilsprotokolle are dedicated word spotting benchmarks and come with a separate set of query word images for QbE and query strings for QbS. There exist 101 string queries for each data set. The query word images in Botany amount to 150 while for Konzilsprotokolle there exist 200.

### 4.2 Word Spotting Protocol

We evaluate our Attribute CNNs in segmentation-based QbE as well as QbS word spotting scenarios. For the data sets GW, IAM DB, Esposalles and IFN/ENIT we follow the protocol proposed in [3] (Almazan Protocol) while for the Botany and Konzilsprotokolle data sets we follow the protocol from the ICHFR 2016 Handwritten Keyword Spotting Competition [37] (Competition Protocol) to be able to directly compare our results to others from the literature.

The Almazan Protocol in [3] is defined as follows: For each data set, the annotation is used to create a segmentation for each word image. For QbE, each word image in the respective test sets is used once as query to rank all the remaining words in the test set which we will refer to as test words under this protocol. Queries that do not have a relevant word among the test words are discarded. For the other queries, however, they are kept as test words in order to act as distractors. For QbS, each unique transcription in the test set is extracted and the respective attribute representations are used as queries to rank all the words in the test set. As each query in QbS has at least one relevant item, no queries are discarded.

The IAM DB is treated differently from the other data sets under the Almazan protocol as for both QbE and QbS stop words are discarded as queries. Again, they are kept as distractors among the test words though.

Different from the Almazan protocol, the segmentation-based Competition Protocol used in [37] makes use of a separate query set. All queries are assumed to have at least one relevant item in the test set thus no query is discarded. Please note that both protocols ignore character cases, e.g. the words "Spotting" and "spotting" are considered to belong to the same class.

---

[5] https://www.prhlt.upv.es/contests/icfhr2016-kws/data.html

In all of our experiments, the attribute representation for a given word image is predicted by the respective Attribute CNN. The query representations are either extracted directly (QbS) or are also obtained from the CNN (QbE). The test words are then ranked based on their Cosine distance to the query representation.

For both word spotting scenarios, we assess the performance of the respective Attribute CNN by calculating the *mean Average Precision (mAP)* which is a standard measure for determining word spotting performance. As the name suggests, the mAP is defined as the mean of the *Average Precision*

$$AP_q = \frac{\sum_{i=1}^{n} P_q(i) r_q(i)}{\text{number of relevant elements}} \quad (15)$$

for each query $q$ where $P_q(i)$ is the precision

$$P = \frac{\text{number of relevant retrieved elements}}{\text{number of retrieved elements}} \quad (16)$$

after cutting off the retrieval list for query $q$ at index $i$. $r_q(i)$ is a function yielding 1 if the $i$-th element of the retrieval list is relevant with respect to $q$ and 0 otherwise. The average precision is equivalent to the area under the interpolated precision-recall curve for a given query.

### 4.3 Creating Attribute Representations

We evaluate three different attribute embeddings: PHOC [3], SPOC [41] and DCToW [64]. Each of the three representations is built using a set of predefined unigrams or alphabet. In our experiments, we determine data-set specific alphabets by extracting all unique characters from the training transcriptions. Thus, attribute representations for data sets with differing alphabets exhibit different dimensionalities.

### 4.4 Training Details

#### 4.4.1 Training Procedure

All the Attribute CNNs used in our experiments are trained in an end-to-end fashion given word images as input and the corresponding attribute representation as label. We do not pre-process the word images but scale their pixels to floating point values in the range of $[0, 1]$ with 0 representing background portions of the word image and 1 representing ink. This is done as we want the CNNs to concentrate on the ink parts of the word image rather than the background.

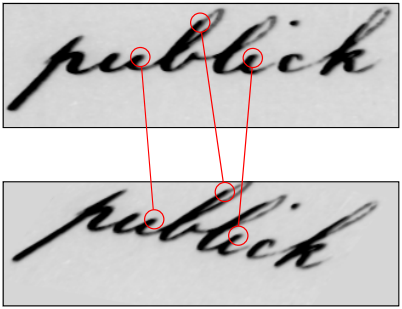

Fig. 5: Visualization of how to extract a synthetic image (bottom) is from an original image (top) for data set augmentation

For training the CNNs to predict PHOCs, we evaluate both Binary Cross Entropy Loss (Eq. 8) and Cosine Loss (Eq. 14). For the other two representations, we use the Cosine Loss only as they cannot be processed by the Binary Cross Entropy Loss.

#### 4.4.2 Regularization

Due to the massive amount of free parameters in the fully connected parts of the PHOCNet and the TPP-PHOCNet, both architectures are prone to overfitting. Hence, we apply a number of regularization techniques which have become standard approaches in deep learning architectures. First, dropout is applied to all but the last fully connected layers (all black layers in Fig. 3). In dropout, the output of a neuron is randomly set to 0 with a given probability. This prevents a neural network to learn certain paths for a given input image "by heart" as neurons can no longer rely on a neuron in a previous layer to always be active for a given image. Another way to think of dropout is the network learning an ensemble of smaller networks which all share weights [56]. The size of this ensemble is exponential in the number of neurons used in the layer applying dropout. For our experiments, we chose a dropout probability of 50%.

In addition to dropout, we augment the number of training images in an unsupervised way. For this, we take three points at fixed relative positions in the middle of an image and multiply each of the coordinates with a random number uniformly sampled from $[0.8; 1.1]$. Then, we compute the homography to obtain the second set of points from the first and use this transformation to generate an augmented image from the original. Fig. 5 illustrates how a new image is generated from an original one. The homography accounts for several transformations that are to be expected in word images at retrieval time, including shear, rotation, translation as well as different slants and scales of the images seen during training.

Another crucial step in training a deep neural network is the weight initialization strategy. As we train our networks with gradient descent, the final solution is inherently dependent on the initial weights of the network. Throughout the literature, various initialization strategies have been used. The influential AlexNet architecture [26], for example, is initialized by drawing weights from a Normal distribution with zero mean and a standard deviation of 0.01. However, choosing initial weights this way hampers training for increasingly deep architectures [16, 53]. In [16], a weight initialization approach is presented which negates the stall in training when using ReLU activation functions in increasingly deeper architectures. We adopt this approach and initialize all weights in our networks from a normal distribution with zero mean and variance $\frac{2}{n}$ where $n$ is the number of inputs for a given filter. As is common for CNNs, the biases in the layers are initialized to zero.

#### 4.4.3 Optimization of the Network Weights

Traditionally, the optimization strategy of choice in deep learning has been stochastic gradient descent with momentum, e.g., [26, 53]. The drawback with this approach is that each weight uses the same learning rate which could possibly hamper training convergence. Recent improvements for classic SGD have thus incorporated additional information into the training process for adapting learning rates individually. For example, AdaGrad [8] assigns low learning rates to frequently occurring features while giving high learning rates to those occurring only rarely. On the other hand, RMSprop [62] normalizes the gradient length for a given weight by a moving average over recent gradient lengths. The optimization strategy Adam [21] combines the advantages of AdaGrad and RMSprop. It works by computing a sliding average for the mean and variance of the gradient for each weight. The weights are then updated by applying the mean gradient normalized with its mean standard deviation. The Adam approach has been increasingly used lately as optimization strategy for deep neural networks, e.g., [12, 20]. In our experiments, we evaluate both standard SGD and Adam optimization.

All our Attribute CNNs are trained with stochastic gradient descent using a mini-batch size of 10. The initial learning rate values are determined by taking the largest value for which training started to converge. For networks being trained with standard SGD this value is $10^{-4}$ when using the Binary Cross Entropy Loss and $10^{-2}$ when using Cosine Loss. For Adam based opti-mization we found that the maximum initial learning rate to achieve convergence is $10^{-4}$ which matches the proposed default value [21]. In order to generate more stable gradients, we use a momentum of 0.9 for the standard SGD. For Adam, we use the recommended hyperparameters $\beta_1 = 0.9$ and $\beta_2 = 0.999$ [21]. We set the weight decay to $5 \cdot 10^{-5}$ as is standard for VGG-style architectures [53]. Training is run for a maximum of 80 000 iterations with the learning rate being divided by 10 after 70 000 iterations. The step size of 70 000 was determined by monitoring the training loss for plateaus. After a total of 80 000 iterations the loss could not be improved upon anymore by lowering the learning rate which is why we stopped training at this point. Only for the experiments on the IAM DB we use a maximum number of iterations of 240 000 as we found the training loss to improve beyond 80 000 iterations. Please note that a training iteration here refers to calculating the gradient for a given mini-batch and updating the weights of the network accordingly.

All the parameters regarding training were chosen based on pre-experiments on the GW data set. Only for the IAM-DB we increased the number of training iterations. Training is carried out on a single Nvidia Pascal P100 using a customized version of the Caffe library [19]. Our Python source code and the custom version of Caffe are made available online[6].

### 4.5 Significance Testing

Before reporting the results of our experiments, we would like to highlight the importance of running statistical significance tests. Unfortunately, it is common practice to simply report mAP values when comparing results of word spotting methods. We advocate for comparing word spotting performances based on statistical significance tests. This allows for assessing whether differences in performance stem from mere chance or are really significant from a statistical point of view.

There exist a number of statistical tests which allow for comparing mAP values. However, most of them make an assumption on the distribution of the test statistic. A notable exception is the permutation test which is also known as resampling or randomization test [54]. We propose to use this significance test whenever comparing mAP results for different word spotting methods.

The null hypothesis for the test is that all average precisions obtained from two different methods (e.g. two different CNNs) on the same data set have equal mean, i.e., the mAP for the two methods is identical.

In order to test this hypothesis, we first compute the observed difference of means, i.e., difference in mAPs. The test then creates random permutations and computes the difference of means of these randomized samples, i.e., the difference of mAPs if average precision values were randomly assigned to one of the two methods. The fraction of permutations where the difference of the randomized average precision values is greater than the observed difference is exactly the $p$-value for the significance test [34, 54]

For an exact test, all possible permutations of the data at hand have to be evaluated. In practice, however, computing all permutations quickly becomes impossible as the sample sizes grow [34]. A solution to this problem is to view the $p$-value of the test as a random variable and approximate it by sampling an adequate number of permutations. Let $\hat{p}$ denote the approximation of the true $p$-value. The standard deviation $s_{\hat{p}}$ of $\hat{p}$ is

$$s_{\hat{p}} = \sqrt{\frac{p(1-p)}{k}} \qquad (17)$$

where $k$ is the number of sampled permutations and $p$ is the underlying true $p$-value [54]. We can rearrange this formula in order to find the number of iterations necessary to obtain a desired standard deviation:

$$k = \frac{p(1-p)}{s_{\hat{p}}^2}. \qquad (18)$$

As the true $p$-value is unknown, we compute the upper limit of permutations necessary to obtain a desired standard deviation for any $p$-value by finding the maximum of $p(1-p)$. The maximum value is obtained for $p = 0.5$ and inserting this into Eq. 18 gives

$$k = \frac{1}{4s_{\hat{p}}^2}. \qquad (19)$$

For our tests, we chose a small desired $s_{\hat{p}}$ of 0.001. Substituting this value into Eq. 19 yields an upper bound of 250 000 random permutations for the permutation test.

### 4.6 Results

Due to the vast amount of possible configurations for our Attribute CNNs, we opt to evaluate the influence of different loss functions and embeddings on the TPP-PHOCNet only and then compare it to the PHOCNet for a smaller amount of feasible set ups. Tab. 2 compares the results obtained for the different string embeddings and loss functions for the TPP-PHOCNet[7].

---

[6] https://github.com/ssudholt/phocnet

[7] We denote the classic stochastic gradient descent optimization as *SGD* and the Adam optimization [21] as *Adam* although technically Adam is a form of stochastic gradient descent as well.

Table 2: Comparison of results using different configurations for the TPP-PHOCNet in mAP [%]. Please note that for values marked with *DNC*, training did not converge to a solution.

| Loss Func. | Embedding | Optimization | GW | | IAM DB | | Esposalles | | IFN/ENIT | |
|---|---|---|---|---|---|---|---|---|---|---|
| | | | QbE | QbS | QbE | QbS | QbE | QbS | QbE | QbS |
| BCE | PHOC | Adam | 97.90 | 96.73 | **84.80** | 92.97 | 97.20 | 94.15 | **96.66** | **94.90** |
| Cosine | PHOC | Adam | 97.15 | 92.35 | 75.11 | 90.27 | 97.23 | 93.64 | 94.06 | 91.38 |
| Cosine | SPOC | Adam | 96.93 | 92.17 | 77.87 | 91.40 | DNC | DNC | 92.31 | 89.46 |
| Cosine | DCToW | Adam | 96.84 | 91.41 | 67.06 | 81.71 | 96.91 | 92.80 | DNC | DNC |
| BCE | PHOC | SGD | 97.75 | 97.50 | 83.38 | 92.59 | 96.93 | **94.33** | 96.49 | 94.68 |
| Cosine | PHOC | SGD | 97.96 | 97.92 | 82.74 | **93.42** | 97.10 | 94.32 | 93.86 | 94.53 |
| Cosine | SPOC | SGD | 97.78 | **98.02** | 82.17 | 92.64 | 97.05 | 94.07 | 94.00 | 94.60 |
| Cosine | DCToW | SGD | **97.98** | 97.65 | 70.44 | 83.02 | 97.11 | 93.75 | 94.40 | 93.81 |

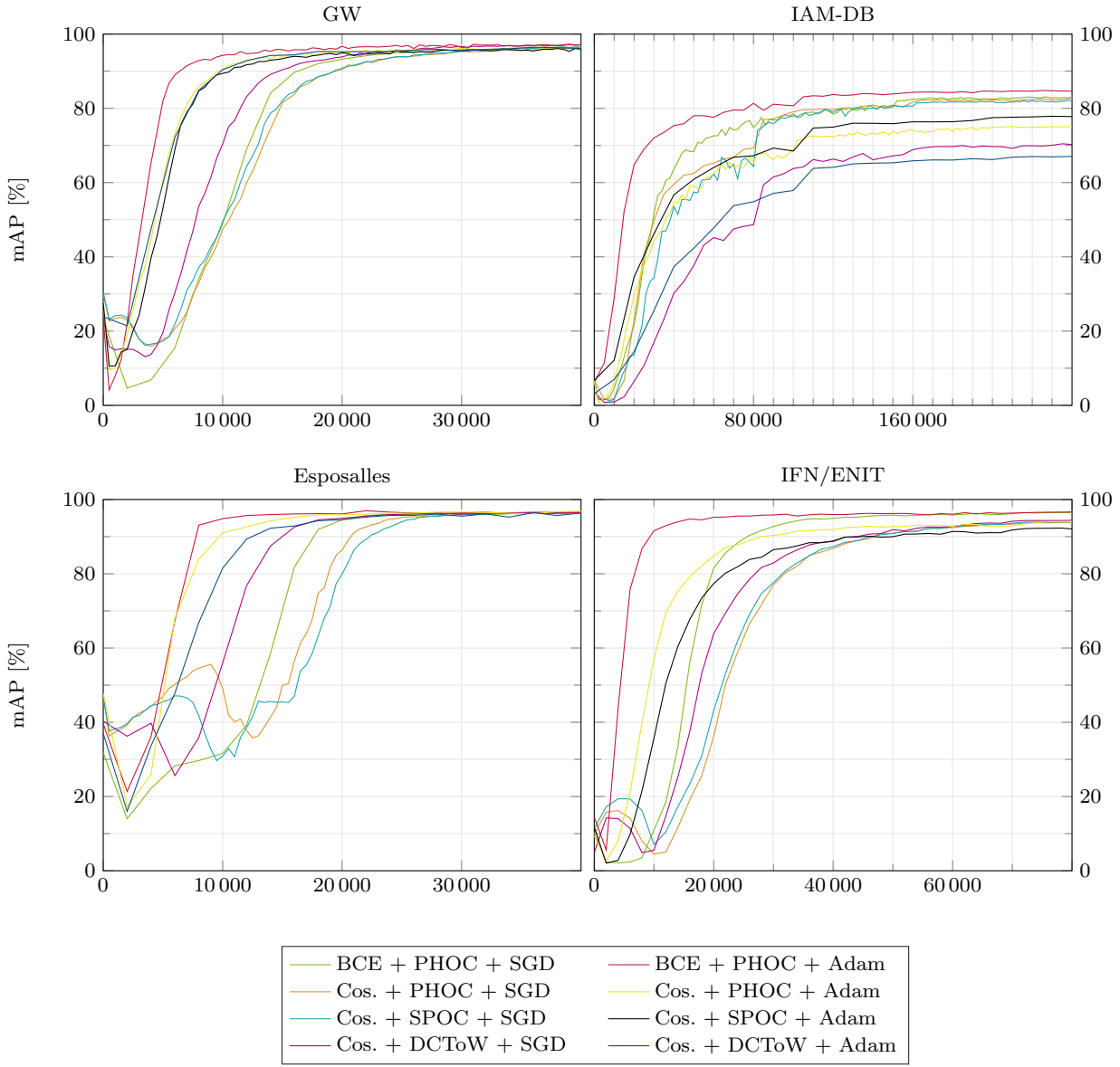

Fig. 6: The figure displays the mAP over the different training iterations for the four QbE experiments evaluated under the Almazan Protocol using the TPP-PHOCNet. Please note that we only show up to 40 000 iterations for the GW and Esposalles experiments as the curves did not exhibit any noticeable difference afterwards.

Table 3: Comparison to results from the literature for experiments run under the Almazan Protocol in mAP [%]

| Method | GW | | IAM | | Esposalles | | IFN/ENIT | |
|---|---|---|---|---|---|---|---|---|
| | QbE | QbS | QbE | QbS | QbE | QbS | QbE | QbS |
| TPP-PHOCNet (BPA) | 97.90 | 96.73 | 84.80 | 92.97 | 97.20 | 94.15 | **96.66** | 94.90 |
| TPP-PHOCNet (CPS) | **97.96** | **97.92** | *82.74* | **93.42** | 97.10 | 94.32 | *93.86* | 94.53 |
| PHOCNet (BPA) | 97.58 | 95.58 | **85.50** | 92.38 | **97.40** | 93.67 | 96.58 | **94.92** |
| PHOCNet (CPS) | 97.72 | 97.44 | *75.85* | **91.12** | 97.17 | **94.89** | *93.33* | 93.87 |
| Deep Feature Embedding [24] | 94.41 | 92.84 | 84.24 | 91.58 | – | – | – | – |
| Attribute SVM + FV [3] | 93.04 | 91.29 | 55.73 | 73.72 | – | – | – | – |
| Finetuned CNN [51] | – | – | 46.53 | – | – | – | – | – |
| LSA Embedding [2] | – | 56.54 | – | – | – | – | – | – |
| Triplet-CNN (*) [64] | 98.00 | 93.69 | 81.58 | 89.49 | – | – | – | – |
| BLSTM (*) [11] | – | 84.00 | – | 78.00 | – | – | – | – |
| SC-HMM (*) [40] | 53.10 | – | – | – | – | - | 41.60 | – |

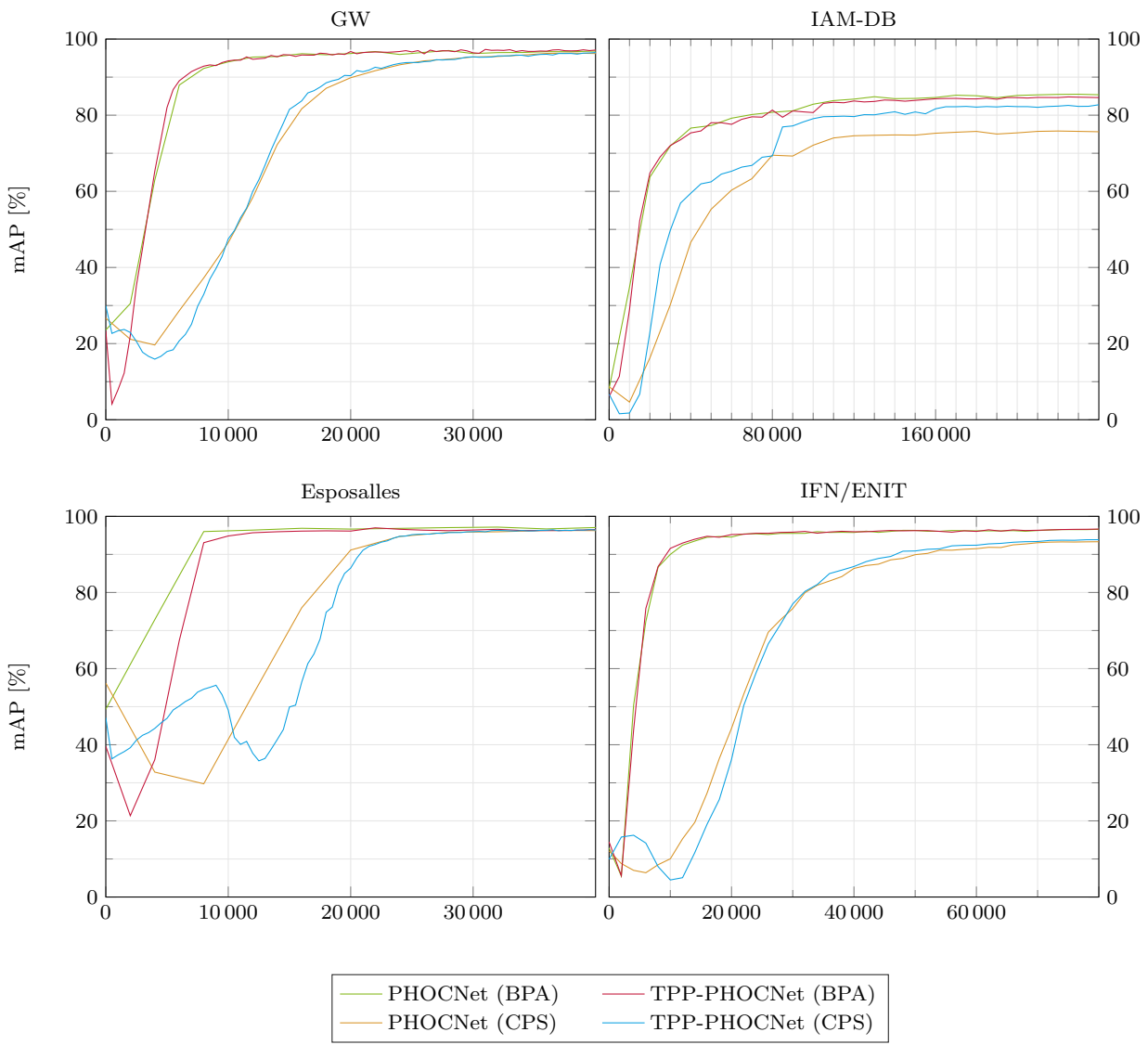

Fig. 7: The figure displays the mAP over the different training iterations for the four QbE experiments using the two different Attribute CNNs.

Table 4: Results for the experiments run on the Botany and Konzilsprotokolle data sets in mAP [%] (results for comparison were obtained from [37])

| Method | Botany Train I | | Botany Train II | | Botany Train III | | Konzilsprotokolle Train I | | Konzilsprotokolle Train II | | Konzilsprotokolle Train III | |
|---|---|---|---|---|---|---|---|---|---|---|---|---|
| | QbE | QbS | QbE | QbS | QbE | QbS | QbE | QbS | QbE | QbS | QbE | QbS |
| TPP-PHOCNet (BPA) | 47.75 | *45.38* | **83.51** | **87.42** | **96.05** | **97.38** | *86.01* | *78.23* | **97.05** | **96.98** | **98.11** | **98.02** |
| TPP-PHOCNet (CPS) | 51.25 | 53.82 | *75.48* | 87.34 | *80.81* | *90.15* | **90.97** | **87.37** | 96.45 | 94.80 | 96.42 | 94.63 |
| PHOCNet (BPA) | *44.56* | *26.62* | 78.93 | *77.22* | 94.10 | 95.43 | *84.34* | *76.45* | 96.05 | 95.27 | 97.08 | 96.22 |
| PHOCNet (CPS) | 45.82 | *42.95* | *71.32* | *81.21* | *79.90* | *89.60* | 88.31 | 83.62 | 95.51 | 93.78 | 95.54 | 93.49 |
| AttributeSVM | **75.77** | **65.69** | – | 65.69 | – | – | 77.91 | 55.27 | – | 82.91 | – | – |
| HOG/LBP | 50.64 | – | – | – | – | – | 71.11 | – | – | – | – | – |
| Triplet-CNN | 54.95 | 3.40 | – | – | – | – | 82.15 | 12.55 | – | – | – | – |

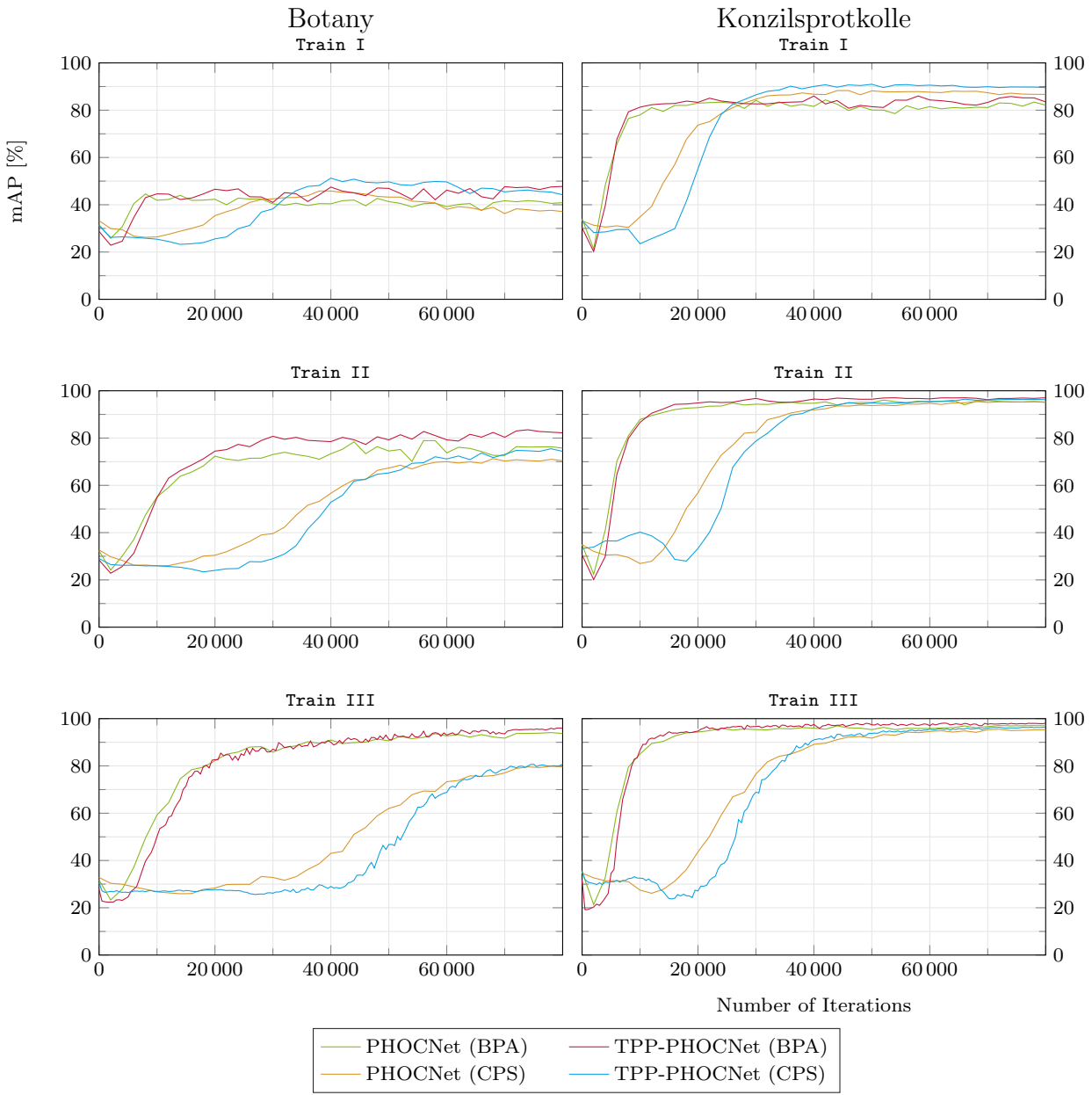

Fig. 8: Comparison of the evolution of the QbE experiments for the Botany and Konzilsprotokolle data sets.

The corresponding Fig. 6 visualizes the evolution of the mAP over training for the four QbE experiments.

Based on the results from Tab. 2, we chose two configurations for which Binary Cross Entropy Loss and Cosine Loss yielded the best results. The first is Binary Cross Entropy Loss, PHOC embedding and Adam optimization (BPA) while the second is Cosine Loss, PHOC embedding and standard SGD optimization (CPS).

Tab. 3 gives a comparison of these configurations for our Attribute CNNs to results obtained from the literature with Fig. 7 showing the corresponding mAP curves. The best results from a numerical point of view are printed in bold. For all regularly printed values no significant difference to the best result can be determined through the permutation test. Finally, all results printed in italics are significantly worse than the best obtained result (significance level $\alpha = 0.05$).

Please note that we can not compare our results to those from the literature through the permutation test as we would need the average precision values for each query instead of the global mAP. Our achieved average precision values will be made publicly available in order for other researchers to compare themselves to our results by means of a statistical test.

Approaches marked with an asterisk do not share the exact same evaluation protocol and can thus not be compared directly to our results. In particular, [40] uses different splits for training and test, effectively reducing the number of test words. This makes the word spotting task easier as the number of distractors is reduced. The BLSTM approach in [11] follows a line spotting protocol. Finally, [64] makes use of the CVL-Database [22] for pre-training their network and thus incorporating more annotated training data.

Finally, Tab. 4 displays the mAP for the experiments evaluated under the competition protocol with Fig. 8 showing the corresponding curves.

4.7 Discussion

4.7.1 Attribute CNN Configurations

The results obtained from the experiments suggest that there is no definitive answer to the question whether one of the word string embeddings examined is superior to another. All embeddings exhibit pathological examples where they perform worse than others. The PHOC embedding might be considered the only exception here as it always achieves state-of-the-art results. In addition, it allows for the fastest training of all configurations when combined with Binary Cross Entropy Loss and Adam optimization (cf. Fig. 6, Fig. 7 and Fig. 8).

The experiments show that replacing the SPP with a TPP layer has a positive influence on the performance for the smaller training partitions of the Botany and Konzilsprotokolle data sets (Tab. 4 and Fig. 8). For the other data sets, both PHOCNet and TPP-PHOCNet achieved similar results. It should be noted, however, that the TPP layer produces a 29% smaller output representation than the SPP layer. This greatly reduces the number of neurons in the ensuing fully connected layer. Furthermore, we found that the TPP layer is able to learn representations that generalize better than the SPP layer. For example, running QbE with the features extracted from the SPP and TPP layers on the IAM DB, the TPP layer is able to achieve 71.41% mAP while the SPP layer only achieves 64.65% mAP. Thus the TPP layer is especially suitable in situations, where a pretrained Attribute CNN is used as a deep feature extractor for document images as is done, e.g., in [39].

4.7.2 Comparison to Results from the Literature

As can be seen in Tab. 3 and Tab. 4 our Attribute CNN architectures achieve state-of-the-art results on all data sets in both QbE and QbS scenarios except for the Train I partition of the Botany data set.

As we do not have the average precision values for other methods, we cannot run the permutation tests in order to assess significance between our results and results from the literature. However, as the TPP-PHOCNet with CPS configuration already achieves a significantly higher mAP than the PHOCNet using the same setup, it is very likely that the mAP for QbS on IAM DB is significantly higher than that of the Deep Feature Embedding and the Triplet-CNN. A similar argument can be made for the QbE scenario on this data set. For the GW, it is likely that the permutation test would not find a significant difference on the mAP values for QbE between our CNNs and the Triplet-CNN as there is already no significant difference between the two mAPs obtained from the TPP-PHOCNets.

4.7.3 Training Set Size Considerations

One of the stigmas that is still attached to CNNs today is that they require large amounts of annotated training. Our Attribute CNNs can be trained with very limited training data from scratch as demonstrated for the GW experiments. Here, the training set encompasses only 3615 annotated samples. As can be seen from the evaluations during training (Fig. 7 and Fig. 8), the regularizations added to our networks prevent any form of overfitting even when faced with as few samples as is the case for the GW data set. We investigated this behavior even further and used as few as 200 training samples and augmented them with synthetic handwritten-like word images from the HW-SYNTH data set [25]. For space limitations, we can unfortunately not discuss all these experiments in this paper. Interested readers may find a detailed report of these experiments in [60]. To summarize the findings: Pretraining a network with synthetic, handwritten-like data allows for drastically reducing the number of training samples necessary to achieve state-of-the-art results. In addition, having the pretrained network, finetuning converges in a manner of minutes, making the approach suitable for human-in-the-loop scenarios.

Table 5: Run times for single forward passes and single PHOC queries [ms]

| Data Set | Forward Pass CPU | Forward Pass GPU | Retrieval |
|---|---|---|---|
| GW | 2191 | 6.2 | 2.0 |
| IAM DB | 3474 | 6.1 | 1.7 |
| Esposalles | 2676 | 5.0 | 1.0 |
| IFN/ENIT | 4309 | 11.5 | 1.2 |
| Botany | 8493 | 15.5 | 0.2 |
| Konzilsprot. | 4318 | 13.8 | 0.2 |

*4.7.4 Run Time Considerations*

In order to assess the applicability of our CNNs in word spotting applications, we investigate the times needed for training and evaluation. When evaluating the times, one has to consider two different points in time: training time and query time. At training time, the system is allowed to be fitted to the data at hand. Run times here are not as critical as query times and can be considered offline precomputations. At query time, however, the user is demanding a responsive system which is able to run the retrieval in a minimal amount of time.

Training times for our Attribute CNNs depend on the number and size of the word images in the data sets. In our experiments, training finished after 9 to 18 hours when run on a Nvidia Pascal P100 GPU.

The query time for our method depends on the time it takes to generate a query representation plus the time for comparing the query representation to the test representations and sorting them. For QbS, the query representation generation time is effectively zero as it can be directly obtained from the string. For QbE, the generation time is the time for the forward pass of the query image. Tab. 5 lists relevant timings for the experiments in milliseconds. Running retrieval with the attribute representations was done using Python and the `sklearn` library[8]. We note that using an advanced graphics card as is done in our case might not be possible in some situations, hence we evaluated timings for a forward pass using a CPU as well. For this we use an Intel Xeon E5-2650 processor which is similar to an Intel Core i7 processor.

From the timings in Tab. 5 it can be seen that a single QbE query takes at most 8.5 s total (forward pass + retrieval) on the CPU and can be as fast as 6 ms when using a GPU. We want to emphasize that the retrieval time for the CPU scenario is almost exclusively due to the forward pass of the query image. In addition, we could increase the corpus size by a factor of 100 in all experiments which would only add at most 200 ms to the query time irrelevant of the usage of a CPU or GPU. Hence, we think that our method is suitable for applications from a timing point of view.

# 5 Conclusion

In this work we present an approach for attribute-based word spotting using CNNs. For this we theoretically derive loss functions which are suitable for training binary as well as real-valued attribute-representations. We are also able to show that the Binary Cross Entropy Loss and the Euclidean Loss yield the same results when the output and the label vectors are normalized.

In addition to the loss functions, we carefully design two AttributeCNN architectures specifically for word spotting. While the first architecture features well-known layer types, the second architecture is equipped with our proposed TPP layer. This layer is able to extract fixed-size representations for arbitrary word images while only considering splits along the horizontal axis.

We evaluate the proposed architectures in a number of experiments on different data sets. The proposed approach is very robust with respect to the choice of meta-parameters.

Though there is no clear cut winner in terms of word string embeddings, we recommend to use the PHOC embedding in conjunction with Binary Cross Entropy Loss and Adam optimization as it did not exhibit one sub-par result in our experiments and achieved the fastest training times.

In this work, we have only focused on segmentation-based word spotting. However, the presented approach can be easily adapted to a segmentation-free scenario as was already shown in [45]. Here, a number of word hypotheses is processed by a TPP-PHOCNet. At query time, the representations for the word hypotheses can be compared to the query representation as is done in the segmentation-based case. As the number of word hypotheses per page are of the same order of magnitude as the largest data sets in the segmentation-based approach, query times per page and per query are similar to the times presented above.

---

[8] http://scikit-learn.org/

**Acknowledgements** We would like to thank Irfan Ahmad for supplying the IFN/ENIT character mapping.